\documentclass[11pt,a4paper]{article}
\usepackage[hyperref]{emnlp2018}
\usepackage{times}
\usepackage{latexsym}
\usepackage{graphicx}
\usepackage{booktabs}
\usepackage{multirow}
\usepackage{amssymb}
\usepackage{url}
\usepackage{multirow}
\usepackage{amsmath}
\usepackage{enumitem}
\usepackage{color}
\usepackage{xcolor,colortbl}
\usepackage{footnote}
\makesavenoteenv{tabular}
\makesavenoteenv{table}

\aclfinalcopy 

\title{
\textsc{Card-660}: 
Cambridge Rare Word Dataset -- a Reliable Benchmark for Infrequent Word Representation Models
}

\author{Mohammad Taher Pilehvar ~~~ Dimitri Kartsaklis ~~~ Victor Prokhorov ~~~ Nigel Collier\\
 Language Technology Lab, 
 Department of Theoretical and Applied Linguistics \\
 University of Cambridge, 
 United Kingdom\\
 {\tt \{mp792,dk426,vp361,nhc30\}@cam.ac.uk}}

\date{}

\begin{document}
\maketitle
\begin{abstract}

Rare word representation has recently enjoyed a surge of interest, owing to the crucial role that effective handling of infrequent words can play in accurate semantic understanding.
However, there is a paucity of reliable benchmarks for evaluation and comparison of these techniques.
We show in this paper that the only existing benchmark (the Stanford Rare Word dataset) suffers from low-confidence annotations and limited vocabulary; hence, it does not constitute a solid comparison framework.
In order to fill this evaluation gap, we propose CAmbridge Rare word Dataset (\textsc{Card-660}), an expert-annotated word similarity dataset which provides a highly reliable, yet challenging, benchmark for rare word representation techniques.
Through a set of experiments we show that even the best mainstream word embeddings, with millions of words in their vocabularies, are unable to achieve performances higher than 0.43 (Pearson correlation) on the dataset, compared to a human-level upperbound of 0.90.
We release the dataset and the annotation materials at \url{https://pilehvar.github.io/card-660/}.


\end{abstract}

\urlstyle{same}

\section{Introduction}

Words in a corpus of natural language utterances approximately follow a Zipfian distribution with their majority, in the ``long tail'' of frequency distribution, occurring rarely. 
The prominent distributional approach to semantic representation relies on enormous occurrences for each individual word; therefore, it falls short of learning accurate representations for rare words in the long tail.
Moreover, it is unreasonable to expect that all words in the vocabulary of a language are observed in a text corpus, even if it is massive in size.
Out-of-vocabulary (OOV) words pose one of the major ongoing challenges for word embedding techniques.
Given that effective handling of rare and OOV words is crucial to accurate natural language understanding, several studies have focused on the topic during the past few years, resulting in a wide range of techniques.

However, despite the popularity of rare and subword semantic representation, the field of research has suffered from the lack of high quality generic evaluation benchmarks.
A task-based evaluation, i.e., one which directly verifies the impact of representation models in a downstream NLP system, despite being very important, does not provide a solid base for comparing different models, given that small variations in the architecture, parameter setting, or initialisation can lead to performance differences.
Moreover, such an evaluation would reflect the ``suitability'' of representations for that specific configuration and for that particular task, and might not be conclusive for other settings.

As far as generic evaluation is concerned, existing benchmarks generally target frequent words.
An exception is the Stanford Rare Word (RW) Similarity dataset \cite{luong-socher-manning:2013} which has been the standard evaluation benchmark for rare word representation techniques for the past few years.
In Section \ref{sec:rw_dataset}, we will provide an in-depth analysis of RW and highlight that crowdsourcing the annotations, with no rigorous checkpoints, has compromised the reliability of the dataset.
This is mainly reflected by the low inter-annotator agreement (IAA), a performance ceiling which is easily surpassed by many existing models.

To overcome this barrier and to fill the gap for a reliable benchmark for the evaluation of subword and rare word representation techniques, we introduce a new dataset, called \textsc{Card-660}: Cambridge Rare Word Dataset.
Compared to existing benchmarks, \textsc{Card-660} provides multiple advantages: (1) thanks to a manual curation by experts, we report IAA of around 0.90 (see Table \ref{tab:iaa}) which is substantially higher than those for existing datasets; (2) word pairs are selected manually from a wide range of domains and, unlike existing datasets, are not bound to a specific resource; (3) word pairs in the dataset are balanced across the similarity scale; and (4) the huge gap between state of the art and IAA (more than 0.50 in terms of Spearman correlation) promises a challenging dataset with lots of potential for future research. 

The paper is structured as follows.
The following Section covers the related work, highlighting some of the issues with the RW dataset.
Section \ref{sec:card-660} details the construction procedure for \textsc{Card-660}.
In Section \ref{sec:analysis},
we analyse the dataset from different aspects, showing how it improves existing benchmarks.
Section \ref{sec:evaluations} reports our evaluation of mainstream word embeddings and recent word representation techniques on the dataset.
Finally, concluding remarks are mentioned in Section \ref{sec:conclusions}.

\section{Related Work}
\label{sec:related_work}

Word similarity datasets have been one of the oldest, still most prominent, benchmarks for the evaluation and comparison of semantic representation techniques.
As a result, several word similarity datasets have been constructed during the past few decades; to name a few: RG-65 \cite{RG65:1965}, WordSim-353 \cite{Finkelsteinetal:2002}, YP-130 \cite{Yang06verbsimilarity}, MEN-3K \cite{Men3k:2014}, SimLex-999 \cite{Hilletal:2015}, and SimVerb-3500 \cite{Gerzetal:2016}.
Many of these English word similarity datasets have been translated to other languages to create frameworks for multilingual \cite{DBLP:journals/corr/LeviantR15} or crosslingual \cite{camachocollados-EtAl:2017:SemEval} semantic representation techniques.
However, these datasets mostly target words that occur frequently in generic texts and, as a result, are not suitable for the evaluation of subword or rare word representation models.

One may opt for transforming a frequent-word benchmarks into an artificial rare word dataset by downsampling the dataset's words in the underlying training corpus
\cite{SergienyaSchutze:2015}.
However, this benchmark might not  properly simulate a real-world rare word representation scenario (cf. Section \ref{sec:motivation}).

\subsection{Stanford RW Dataset}
\label{sec:rw_dataset}

The Stanford Rare Word Similarity (RW) dataset is an exception as it is dedicated to evaluating infrequent word representations. 
The dataset has been regarded as the de facto standard evaluation benchmark for subword and rare word representation techniques.
However, our analysis shows that RW suffers from multiple issues: (1) skewed distribution of the scores, (2)
low-quality and inconsistent scores, and as a consequence, (3) low inter-annotator agreement.

The RW dataset comprises 2034 word pairs (i.e., \textit{word$_1$} -- \textit{word$_2$}).
Candidates for \textit{word$_1$} were randomly sampled from Wikipedia documents, distributed across a wide range of frequencies (from 5 to 10,000) to ensure the inclusion of infrequent words.
Given this automatic sampling, a measure was required to avoid noisy or junk words.
To this end, a sampled word was checked in WordNet \cite{Fellbaum:98} and was included only if it appeared in at least one synset.
Hence, the vocabulary of the dataset is bound to that of WordNet.
Words for \textit{word$_2$} were randomly picked from synsets that were directly connected to a synset of \textit{word$_1$}, through various relations, such as hypernymy, holonymy, and attributes.

\begin{figure}[t!]
\begin{center}
	\includegraphics[trim = 0mm 25mm 0mm 0mm,scale=0.21]{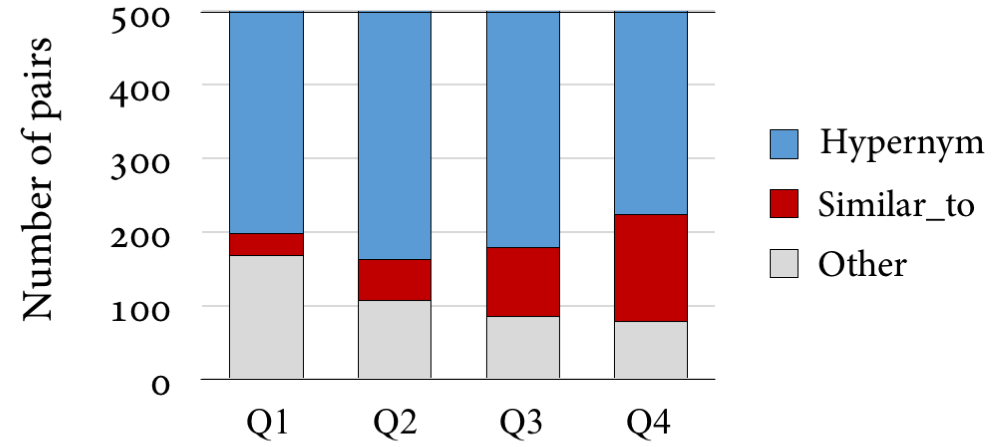}
\end{center}
 \caption{Distribution of relation types (``hypernymy'', ``similar\_to'', and others) across four quartiles (sorted by gold similarity scores) of the Stanford Rare Word Similarity dataset. The distribution of pairs with hypernymy relation is almost uniform across the quartiles, whereas one would expect many more pairs in the top quartiles (Q4 and Q3), given the high semantic similarity of hypernym-hyponyms. 
 }
 \label{fig:rw_dataset}
\end{figure}

\subsubsection{Distribution of scores}
These word pairs were assigned similarity scores in $[0,10]$.
Given that all word pairs in the dataset are semantically-related according to WordNet, the scores form a skewed distribution biased towards the upper bound (see Figure \ref{fig:distribution} and Section \ref{sec:score-distribution} for more details).

\subsubsection{Consistency of annotations}
The scoring of the pairs has been carried out through crowdsourcing: (Amazon Mechanical) Turkers have provided ten scores for each word pair.
The raters were restricted to only US-based workers and they were asked to self-certify themselves by indicating if they ``knew'' the word; this was used to ``discard unreliable pairs.''
However, our analysis of the dataset clearly indicates that the above measures have not been adequate for guaranteeing quality annotations.
For instance, the word \textit{bluejacket} is paired with \textit{submariner} in the dataset. 
According to WordNet (v3.0), a submariner (``a member of the crew of a submarine'') is a bluejacket (``a serviceman in the navy''; a \textit{navy\_man}, \textit{sailor}), hence a hypernymy relationship.
One would expect a word to have high semantic similarity with its hypernym.
However, the gold score for this pair is just 0.43 in the scale $[0,10]$.
Other examples include ``untruth'' (a false statement) vs. ``statement'' (again, with a hypernymy relationship) with a low similarity of 1.22.
Apart from not being a rigorous evaluation, the self-certification does not verify if the annotator had knowledge of \textit{various} possible meanings of a word. 
For instance, \textit{decomposition} could refer to the analysis of vectors in algebra; but, when paired with \textit{algebra}, the assigned score is only 0.75.
Such examples clearly indicate that the annotators were not aware of specialised senses of some words (e.g., the algebraic meaning of decomposition), despite ``knowing'' the word.

In fact, there are numerous such pairs in the dataset.
According to our estimate, 78\% of the 2034 word pairs in the dataset are in a \textit{hypernymy} or \textit{similar\_to} relationship.
One would expect most of these (semantically similar) pairs to have been assigned high similarity scores which are closer to the upper bound of the similarity scale $[0,10]$.
However, as shown in Figure \ref{fig:rw_dataset}, these pairs are spread across the similarity scale, spanning from complete unrelatedness (lower bound) to identical semantics (synonymy).
Having the words in the dataset sorted by their assigned gold scores, respectively, 66\%, 79\%, 83\%, and 85\% of the pairs in the first to fourth quartiles contain either ``hypernymy'' or ``similar\_to'' relations (whereas one would expect most of these semantically-similar pairs to appear in the top quartiles).


Additionally, the dataset suffers from inconsistent annotations. For instance, the two almost identical pairs \textit{tricolour}-\textit{flag} and \textit{tricolor}-\textit{flag} were assigned substantially different scores, i.e., 5.80 and 0.71, respectively.
This inconsistency is also reflected by high variances across annotators’ scores (cf. Section \ref{sec:consistency}).

\subsubsection{Inter-Annotator Agreement (IAA)} 
This validity metric reflects the homogeneity of the annotators' ratings and it is generally accepted as the upper bound for machine performance.
IAA is widely used as a standard evaluation metric for the quality of word similarity datasets. 
A low IAA indicates a defective similarity scale or unreliable annotations.

In the RW dataset, ``up to 10'' annotations have been provided for the 2034 word pairs, each with a similarity score in $[0,10]$ range. 
More precisely, 214 of the pairs are not provided with 10 scores, with the minimum number of scores for a pair being 7.
The authors did not report IAA statistics for this dataset. Given that the annotators are not known for each pair in the released dataset, it is not straightforward to compute IAA.\footnote{The scores are further pruned down to only those that were within one standard deviation of the mean. This results in a further imbalanced set of scores, making the computation of IAA more challenging.}
According to a rough calculation, the average pairwise Spearman correlation between annotators' scores is 0.40, which is a significantly low figure compared to other existing word similarity datasets. We report an impressive IAA of 0.89 for our dataset (cf. Section \ref{sec:iaa}).

\section{The \textsc{Card-660} Dataset}
\label{sec:card-660}

\subsection{Motivation}
\label{sec:motivation}

Due to a lack of reliable evaluation benchmarks, research in rare word representation has often resorted to artificial experimental setups such as corpus downsampling \cite{SergienyaSchutze:2015,herbelot-baroni:2017:EMNLP2017,Angeliki2017-LAZMWM}. 
To this end, in order to simulate a rare word scenario, the rare word representation model is provided with only a limited number of occurrences for the target set of words, for instance by means of replacing the dataset's words with some other sequences of characters (e.g., by augmenting ``UNK'', such as ``skyglowUNK'' for ``skyglow'') in the training corpus.
The computed representations on the ``downsampled'' training data are then either evaluated on a standard word similarity dataset \cite{SergienyaSchutze:2015}, such as RG-65, or compared against reference embeddings computed on a large training corpus \cite{herbelot-baroni:2017:EMNLP2017,Angeliki2017-LAZMWM}.

However, due to the following three reasons, downsampling does not constitute a reliable benchmark that can represent the challenging nature of the task: 
(1) it is unable to control the impact of second-order associations (words that frequently co-occur with the downsampled word) and cannot represent a real-world setting with novel rare usages; (2) given that morphological variations of a word (such as plural forms) are kept intact in this procedure, a subword technique can easily resort to these forms to compute the embedding for downsampled words; and (3) a constrained evaluation configuration in which the task is to estimate the embedding for a (rare) word using one or few occurrences (contexts) of it, limits the benchmark to a subset of corpus-based rare word representation techniques only.
Moreover, the evaluation would require the comparison of the computed embeddings for rare words with a set of reference embeddings (computed on the full data).
This dependency limits the ability of the benchmark in providing a direct evaluation of the rare word representation technique, independently from the impact of the model used to compute the reference embeddings.

The \textsc{Card-660} dataset aims at filling the gap for rigorous generic evaluation of rare word and subword representation models.
In what follows in this section, we will detail the construction procedure of the dataset which was carefully planned to guarantee a challenging and reliable dataset.

\begin{table*}[t!]
  \centering
  \setlength{\tabcolsep}{12pt}
  \renewcommand{\arraystretch}{1.2}
  \footnotesize
  \begin{tabular}{c l l}
    \toprule
    \bf Score & \bf Interpretation & \bf Example pair\\
    \midrule
    \bf 4 & {\bf Synonyms.} The two words are different ways of referring to the same concept &  \textit{car} ~~~ \textit{automobile} \\

    \bf 3 & {\bf Similar.} The two words are of the same nature, but slightly different in details & \textit{car}~~~\textit{truck} 
    \\

    \bf 2 & {\bf Related.} The two words are closely related but they are not similar in their nature & \textit{car}~~~\textit{driver}\\

    \bf 1 & {\bf Same domain or slight relation.} The two words have distant relationship & \textit{car}~~~\textit{tarmac} \\

    \bf 0 & {\bf Completely unrelated.} The two words have nothing in common.
    & \textit{car}~~~\textit{sky} \\
    \bottomrule    
  \end{tabular}
  \caption{The five-point Likert similarity scale used for the annotation of the dataset.}
  \label{tab:rating-scale}
\end{table*}

\begin{table}[t!]
  \centering
  \renewcommand{\arraystretch}{1.3}
  \small
 {
  \begin{tabular}{p{0.45\textwidth}}
    \toprule
    {\bf Task}. Resource \\
    \midrule
    {\bf Text classification.} BBC \cite{greene06icml} \\
    {\bf Sentiment analysis.}  IMDB \cite{maas-EtAl:2011:ACL-HLT2011}, Multi-Domain Sentiment Dataset \cite{P07-1056} \\  
    
    {\bf Machine Translation.} Europarl \cite{koehn2005epc}    \\
    {\bf Question Answering.}  AQUA-RAT \cite{ling-EtAl:2017:Long}, SQuAD \cite{rajpurkar-EtAl:2016:EMNLP2016}\\
    {\bf BioMedical (entity recognition).}    JNLPBA corpus \cite{Kim:2004:IBR} \\
    {\bf Social media.}  Twitter\\
    {\bf Ontologies and online glossaries.}  WordNet, Wiktionary  \\
    {\bf Named entities.} Freebase \cite{Bollackeretal:2008} \\
    
    {\bf Veracity assessment.}   FakeNews\footnote{\url{http://www.fakenewschallenge.org}}\\

    \bottomrule    
  \end{tabular}}
  \caption{Various datasets and resources used for rare word selection in \textsc{Card-660}.}
  \label{tab:resources}
\end{table}

\subsection{Construction Procedure}

The following four-phase procedure was used to construct the dataset: 
\begin{itemize}[noitemsep]
    \item[(1)] A set of 660 rare words were carefully selected from a wide range of domains; 
    \item[(2)] For each of these initial words, a pairing word was manually selected according to a randomly sampled score from the similarity scale (Section \ref{sec:word_selection}); 
    \item[(3)] All pairs were scored by 8 annotators; 
    \item[(4)] A final adjudication was performed to address disagreements (Section \ref{sec:scoring}).
\end{itemize}



\subsubsection{Similarity scale}
\label{sec:sim-scale}

We adopted the five-point Likert scale used for the annotation of the datasets in SemEval-2017 Task 2 \cite{camachocollados-EtAl:2017:SemEval}.
The task reported high IAA scores which reflects the well-definedness and clarity of the scale.
We provided annotators with the concise guideline shown in Table \ref{tab:rating-scale}, along with several examples.
Given the continuity of the scale, the annotators were given flexibility to select values in between the five points, whenever appropriate, with a step size of 0.5.

The annotators were asked in the guidelines to make sure they were familiar with \textit{all} common meanings of the word (as defined by WordNet or other online dictionaries).
To facilitate the annotation, the annotators were provided with the definitions of some of the words that were defined in WordNet or named entities that had Wikipedia pages.
For others, we asked the annotators to check the word in online dictionaries, such as WordNet browser\footnote{\url{http://wordnetweb.princeton.edu/perl/webwn}} and Wiktionary\footnote{\url{www.wiktionary.org}}, or encyclopediae, such as Wikipedia.

\subsubsection{Word pair selection}
\label{sec:word_selection}

Unlike previous work \cite{luong-socher-manning:2013},
we did not rely on random sampling (pruned by frequency) of initial words from a specific dictionary, to prevent the dataset from being restricted to a specific resource or vocabulary.
Instead, we carefully hand-picked word pairs from a wide range of domains.
To construct the 660 pairs of the dataset (each pair is denoted as $w_1-w_2$), we first picked 660 $w_1$ words.
Our aim was to have a dataset that can ideally reflect the performance of rare word representation techniques in downstream NLP tasks.
To this end, we picked initial words ($w_1$s) from different common NLP datasets and resources, listed in Table \ref{tab:resources}.
For each text-based resource, a frequency list was obtained and rare words were carefully picked from the long tail of the list, cross-checking the frequency of words in the Google News dataset.
For the other resources (such as Wiktionary), we checked a word against a large frequency list to ensure they are not frequent words. 
The list was computed on the 2.8B token ukWaC+WaCkypedia corpus \cite{Baroni2009Wacky} and comprised 16.5M unique words.

In order to have a balanced distribution of scores in the dataset, we first assigned random integer scores in $[0-4]$ to the 660 initial $w_1$s.
Then, with the corresponding score in mind, a pairing word ($w_2$) was selected for each $w_1$.
We show in Section \ref{sec:score-distribution} that this strategy resulted in a uniformly distributed set of scores in the dataset.

The dataset comprises words from a wide range of genres and domains, including slang in social media (e.g., \textit{2mrw} and \textit{Mnhttn}), named entities (e.g., \textit{Stephen\_Hawking} and \textit{Ursa\_Major}), and domain specific terms (e.g., \textit{erythroleukemia} and \textit{NetMeeting}).
Moreover, to have a rigorous testbed for subword representation techniques that emphasises the importance of semantic (rather than shallow) understanding of the words, the dataset contains several word pairs that have similar surface forms (hence, high string similarity) while being semantically distant, e.g., \textit{infection}-\textit{inflection} and \textit{currency}-\textit{concurrency}.
There are also many compound words (e.g., \textit{skyglow}, \textit{musclebike}, and \textit{logboat}) which makes the dataset particularly interesting for evaluating compositionality as well as for subword representation techniques.

\subsubsection{Scoring and adjudication}
\label{sec:scoring}

Once the 660 word pairs were manually selected (by the first author), the initial scores were discarded and the words were shuffled (vertically and horizontally) to dispense any potential bias from the initial round of creation.
Then, the pairs were assigned to 8 annotators (including all but first authors) who independently scored each and every pair according to the annotation guidelines (see Section \ref{sec:sim-scale}).
All annotators were PhD graduates or students in Computational Linguistics or related fields and were either native or fluent English speakers.

Once all pairs were scored by the annotators, we checked for disagreements.
This check was intended to improve the dataset's quality through resolving simple annotation mistakes.
For each annotator, we marked the $i^{th}$ pair if for the assigned score $s_i$: $s_i \ge \mu_i+1$ or $s_i \le \mu_i-1$, where $\mu_i$ is the average of the other seven annotators' scores for $s_i$.
The annotator was then asked to (more carefully) re-score the marked pair by checking for its possible meanings.
They were asked to keep their initial score if not convinced otherwise.
The adjudication revealed that most disagreements were due to an annotator having misread a word or not been familiar with a specific meaning of it, or missing annotations.
By average, 13.8\% of the pairs were re-scored by each annotator.

\section{Analysis}
\label{sec:analysis}

In this section we provide an analysis on the quality of \textsc{Card-660} from three different perspectives: distribution of scores, inter-annotator agreement, and consistency among annotators.
We benchmark \textsc{Card-660} against the Stanford {\bf RW} dataset and two standard word similarity datasets (cf. Section \ref{sec:related_work}): SimVerb-3500 (\textbf{SV-3500}) and SimLex-999 (\textbf{SL-999}).
The latter two datasets do not target rare words; however, given that their construction strategy is similar to that employed for creating RW (based on crowdsourcing), we included them in our analysis experiments to provide better insights.
For the purpose of this evaluation, all the datasets were scaled to [0,10] to make them comparable.

\subsection{Score Distribution}
\label{sec:score-distribution}

\begin{figure}[t!]
\begin{center}
	\includegraphics[trim = 0mm 15mm 0mm 0mm, scale=0.125]{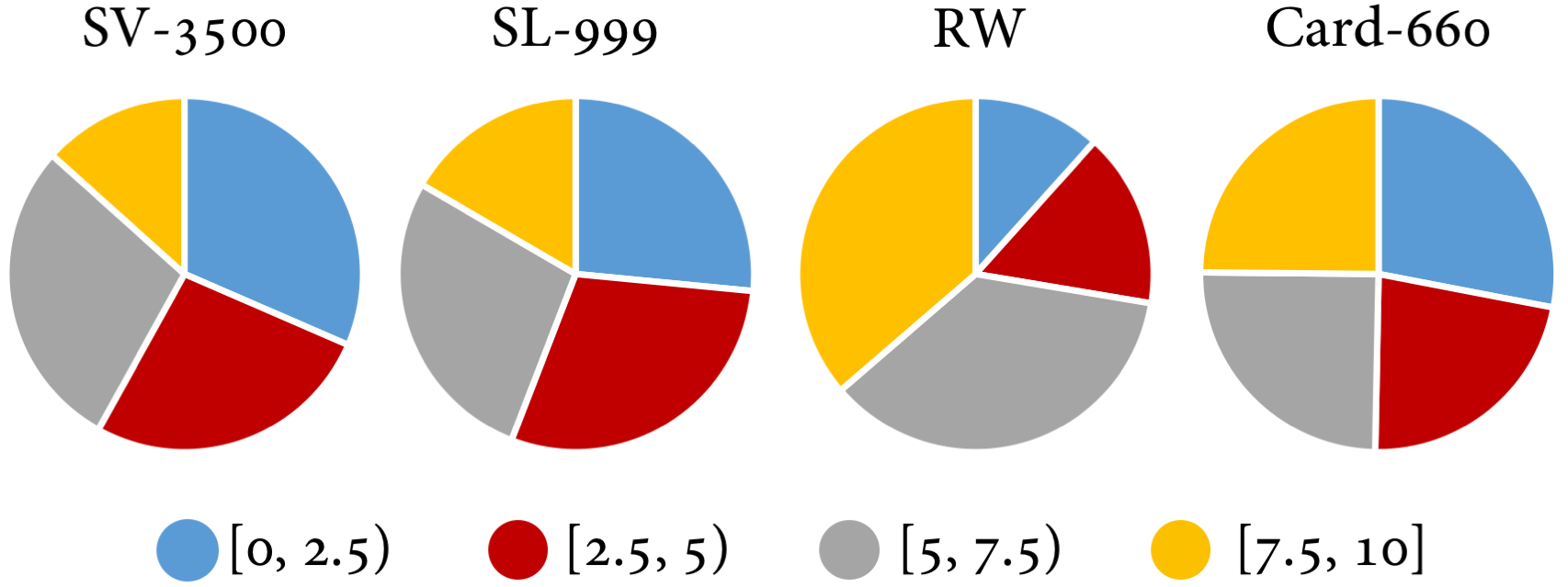}
\end{center}
 \caption{The distribution of word pairs across the four quartiles of the similarity scale for different datasets. A perfectly balanced dataset would have four equally sized slices.}
 \label{fig:distribution}
\end{figure}

Figure \ref{fig:distribution} shows the distribution of pairs across the similarity scale, for \textsc{Card-660} and the three other datasets.
As discussed in Section \ref{sec:rw_dataset}, RW is heavily biased towards the upper bound of the similarity scale (with around 72\% of the pairs in the upper half, i.e., $[5,10]$).
The skewed distribution in this dataset can be attributed to the automatic word pair selection from semantically-related words in WordNet (cf. Section \ref{sec:rw_dataset}).
SV-3500 and SL-999 are skewed towards the lower bound, but to a smaller degree (around 59\% of the pairs in $[0,5)$).
Thanks to the manual creation of \textsc{Card-660}, we have a balanced set of pairs across the similarity scale (50-50\% across the two halves).

\begin{figure}[t!]
\begin{center}
	\includegraphics[trim = 0mm 15mm 0mm -10mm,scale=0.18]{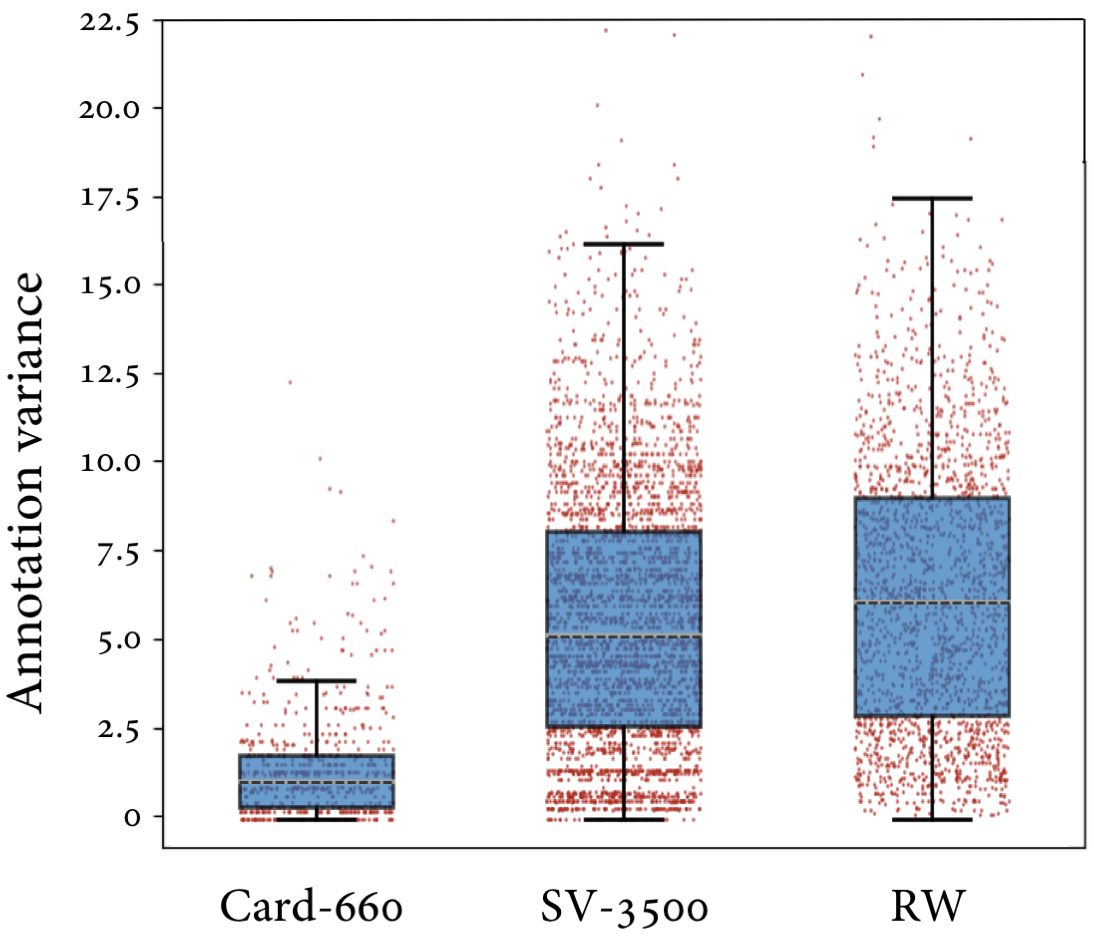}
\end{center}
 \caption{Annotation variance for word pairs across different datasets. Average variance for \textsc{Card-660} is 1.47, which is significantly lower than those for SV-3500 and RW: 5.64 and 6.34, respectively. 
 }
 \label{fig:variance}
\end{figure}

\subsection{Inter-Annotator Agreement}
\label{sec:iaa}

As mentioned in Section \ref{sec:rw_dataset}, IAA has been extensively used as a quality metric for word similarity datasets.
Following standard practise, we measure two sets of IAA scores: (1) {\bf Pairwise} is the averaged pairwise correlation between all possible rater pairings, and (2) {\bf Mean} is the averaged correlation of each rater against the average of others. 

Table \ref{tab:iaa} reports IAA statistics for \textsc{Card-660}.
Thanks to the manual scoring of the pairs by experts (as opposed to turkers), the IAA values for the dataset
are very high, placing it among the best word similarity datasets in the literature.
This is particularly interesting considering that, compared to standard word similarity datasets which contain mostly common words, our dataset comprises words that are semantically difficult to annotate due to their rare nature.
The pairwise IAA score of 88.9 is significantly higher than the crowdsourced RW, with the estimated pairwise IAA score of around 40.0 (cf. Section \ref{sec:rw_dataset}).
The same applies to other recent crowdsourced word similarity datasets for common words which usually report pairwise IAA scores below 70.0 (e.g., $\rho=67.0$ for SL-999)\footnote{SimVerb-3500 reports a pairwise $\rho$ of 84.0; however, our calculation did not agree with this figure. Personal communication with the authors revealed an issue in the computation of their IAA. The correct figure is instead 61.2.}.


\begin{table}[t!]
  \centering
  \setlength{\tabcolsep}{5pt}
 \scalebox{0.9}
 {
  \begin{tabular}{l c c c c}
    \toprule
      & \multicolumn{2}{c}{\bf Mean} & \multicolumn{2}{c}{\bf Pairwise} \\
      \cmidrule(lr){2-3}
      \cmidrule(lr){4-5}
      & $r$ &   $\rho$  &   $r$ &   $\rho$  \\
      \midrule
Initial &   88.0$\pm$2.3    &   87.9$\pm$1.9    &   80.2$\pm$2.9    &   80.6$\pm$2.6    \\
Final &   93.5$\pm$1.4    &   93.1$\pm$1.2 &
88.9$\pm$1.7 & 88.9$\pm$1.7 \\
    \bottomrule
  \end{tabular}}
  \caption{Inter-annotator agreement (IAA) scores
  before (\textit{initial}) and after (\textit{final}) adjudication ($\pm$ standard deviation). 
  IAA is shown in terms of Pearson $r$ and Spearman $\rho$ percentage correlations. The \textit{final} scores are representative of the dataset's quality.}
  \label{tab:iaa}
\end{table}

\definecolor{Gray}{gray}{0.95}
\newcolumntype{a}{>{\columncolor{Gray}}c}

\begin{table*}[t!]
  \centering
  \setlength{\tabcolsep}{4pt}
 \scalebox{0.9}
 {
   \begin{tabular}{l r| ra ra| c ra ra|}
    \toprule
    \multirow{2}{*}{\bf Embedding set} &  
    \multicolumn{1}{c}{\multirow{2}{*}{\bf $|V|$}} &
    \multicolumn{2}{c}{\bf Missed words} & 
    \multicolumn{2}{c}{\bf Missed pairs} & &
    \multicolumn{2}{c}{\bf Pearson $r$} &   
    \multicolumn{2}{c}{\bf Spearman $\rho$}  \\
    \cmidrule(lr){3-4}
    \cmidrule(lr){5-6}
    \cmidrule(lr){8-9}
    \cmidrule(lr){10-11}
    & \multicolumn{1}{c}{} & RW & \textsc{Card} & RW & \multicolumn{1}{c}{\textsc{Card}} & & RW & \textsc{Card} & RW & \multicolumn{1}{c}{\textsc{Card}} \\
    
    \midrule
    Glove Wikipedia-Gigaword (300d)  &   400K    & ~~7\% & 55\%  & 12\% & 74\%  && 34.9 & 15.1 & 34.4 & 15.7 \\
    Glove Common Crawl - uncased (300d)   &   1.9M  &  1\% & 36\% & 1\% & 50\% && 36.5 & 29.2 & 37.7 & 27.6\\
    Glove Common Crawl - cased (300d)   &   2.2M    & 1\% & 29\% & 2\% & 44\%  && 44.0 & 33.0 & 45.1 & 27.3\\
    Glove Twitter (200d)   &   1.2M   & 29\% & 60\%  & 48\% & 79\%  &&  17.7 & 13.7 & 15.3 & 11.7\\
    Word2vec GoogleNews (300d) &   ~~~3M  & ~~6\%& 48\% & 10\% & 75\% && 43.8 & 13.5 & 45.3 & ~~7.4 \\ 
    Word2vec Freebase (1000d) &   1.4M  & 100\% & 85\% & 100\% & 92\%  && 0.0 & 17.3 & 0.0 & ~~4.6 \\ 
    Dependency-based Wikipedia (300d) &   174K   & 22\% & 60\% & 36\% & 80\% && 17.4 & ~~6.4 & 19.7 & ~~3.3 \\
    LexVec Common Crawl (300d) &   ~~~2M  & 1\% & 41\% &  1\% & 55\%  && 47.1 & 25.9  & 48.8 &  18.5 \\
    LexVec Wikipedia-NewsCrawl (300d)  &   370K    & 8\% & 58\% & 14\% &  78\% && 35.6 & 11.8 & 34.8 & ~~7.8  \\
    ConceptNet Numberbatch (300d) &   417K  & ~~5\% &  37\%  & 10\%& 53\%  && 53.0 & 36.0 & 53.7 & 24.7  \\
    \midrule
    ConceptNet + Word2vec Freebase &   1.6M  &  1\% & 22\%  & 2\% & 45\%  && 44.0 & 42.6 & 45.1 & 31.3  \\
    Glove cased CC + Word2vec Freebase &   3.4M  &  11\% & 21\% & 10\% & 39\%  && 53.0 & 38.8 & 53.7 & 32.7  \\

    \bottomrule
  \end{tabular}}
  \caption{Pearson $r$ and Spearman $\rho$ correlation percentage performance of mainstream pre-trained word embeddings on the RW and \textsc{Card-660} datasets. Column $|V|$ shows the size of vocabulary for the corresponding embedding set.}
  \label{tab:word_embeddings}
\end{table*} 

\subsection{Consistency of Annotations}
\label{sec:consistency}

Despite being suitable for measuring linear relationships between scores, correlation cannot fully reflect the consistency between annotators.
Two annotators can have perfect correlation, i.e., 1.0, even if they consistently provide different scores for the same pairs (therefore, having different average assigned scores).
To check the consistency among annotators, i.e., if they had the same interpretation of the similarity scale, we compute variance across annotators for individual pairs.

The box and whisker (over scatter) plot in Figure \ref{fig:variance} shows the distribution of annotator variances for the pairs in different datasets.
Clearly, the score variances for \textsc{Card-660} are significantly lower than those for the two crowdsourced datasets, i.e., SimVerb-3500 and RW.\footnote{We are not able to report results for SimLex-999 since individual annotators' scores are not released for this dataset.}
Specifically, for the majority of pairs in \textsc{Card-660} the annotation variance is lower than the other two datasets' first quartile (bottom of the blue square which splits the lower 25\% of the data from the top 75\%). 
This indicates that our annotators had significantly higher degrees of agreement, reflecting the well-definedness of the similarity scale as well as the reliability of expert-based annotation (as opposed to crowdsourcing).


\section{Evaluations}
\label{sec:evaluations}

In the remainder of this paper, we provide two sets of experiments to showcase the challenging nature of our dataset.
Specifically, in Section \ref{sec:pretrained} we report the performance of common pretrained word embeddings on \textsc{Card-660}, and in Section \ref{sec:rare_rep} we provide experimental results for state-of-the-art rare word representation techniques.
In all experiments, we used the cosine similarity for comparing pairs of word embeddings.

\subsection{Pre-trained Embeddings}
\label{sec:pretrained}

As was mentioned earlier in the Introduction, it is not possible to enumerate the entire vocabulary of a natural language, even if massive corpora are used.
A challenging rare word benchmark should ideally reflect this phenomenon.
To verify this in our dataset, we experimented with a set of commonly used word embeddings trained on corpora with billions of tokens.

Table \ref{tab:word_embeddings} provides correlation performance results for different embedding sets on the RW and \textsc{Card-660} datasets.
Specifically, we considered different variants of 
Word2vec\footnote{\url{https://code.google.com/archive/p/word2vec/}} \cite{Mikolovetal:2013} and Glove\footnote{\url{https://nlp.stanford.edu/projects/glove/}} \cite{Penningtonetal:2014}, two commonly-used word embeddings that are trained on massively large text corpora; Dependency-based embeddings\footnote{\url{http://u.cs.biu.ac.il/~yogo/data/syntemb/deps.words.bz2}} \cite{levy-goldberg:2014:P14-2} which extends the Skip-gram model to handle dependency-based contexts; LexVec\footnote{\url{https://github.com/alexandres/lexvec}} \cite{salle-villavicencio-idiart:2016:P16-2} which improves the Skip-gram model to better handle frequent words; and ConceptNet Numberbatch\footnote{\url{https://github.com/commonsense/conceptnet-numberbatch}} \cite{speer2017conceptnet} which exploits lexical knowledge from multiple resources, such as Wiktionary and WordNet, and was the best performing system in SemEval 2017 Task 2.
In the last two rows of the Table we also report results for two hybrid embeddings constructed by combining the pre-trained Freebase Word2vec, which mostly comprises named entities, with two of the best performing embeddings evaluated on the dataset.
Given that the word embeddings are not comparable across two different spaces, we only compute the similarity between a pair only if both words are covered in the same space (with priority given to the non-Freebase embedding).

As can be seen in the Table, many of the embeddings yield high coverage for the RW dataset, with those trained on the Common Crawl (CC) corpus providing near full coverage.
This highlights the limited vocabulary of the dataset (which is bound to that of WordNet).
Also, many of the embeddings attain performance around 40.0 on RW, which is higher than the estimated IAA of the dataset.
In contrast, \textsc{Card-660} proves to be significantly more challenging, with the highest coverage model (Glove CC and Word2vec Freebase hybrid model) missing around 40\% of the pairs.
Also, the best performance of 42.6 ($r$) and 32.7 ($\rho$) are substantially (around 50.0) lower than the IAA for the dataset (see Table \ref{tab:iaa}).

\begin{table*}[t!]
  \centering
  \setlength{\tabcolsep}{3.6pt}
 \scalebox{0.9}
 {
  \begin{tabular}{c l ra ra| c ra ra|}
    \toprule
    \multicolumn{2}{c}{\multirow{2}{*}{\bf Model}} &  
    \multicolumn{2}{c}{\bf Missed words} & 
    \multicolumn{2}{c}{\bf Missed pairs} &  &
    \multicolumn{2}{c}{\bf Pearson $r$} &   
    \multicolumn{2}{c}{\bf Spearman $\rho$}  \\
    \cmidrule(lr){3-4}
    \cmidrule(lr){5-6}
    \cmidrule(lr){8-9}
    \cmidrule(lr){10-11}
    & & RW & \textsc{Card} & RW & \multicolumn{1}{c}{\textsc{Card}} & & RW & \textsc{Card} & RW & \multicolumn{1}{c}{\textsc{Card}} \\
    \midrule
    \multicolumn{2}{l}{\textit{ConceptNet Numberbatch (300d)}} & ~~5\% &  37\%  & 10\%& 53\%  & & 53.0 & 36.0 & 53.7 & 24.7  \\
    & + Mimick \cite{pinter-guthrie-eisenstein:2017:EMNLP2017}  &  0\%  & ~~0\%  & ~~0\%  & ~~0\% & & 56.0 & 34.2 & 57.6 & \bf \underline{35.6} \\
    & + Definition centroid \cite{herbelot-baroni:2017:EMNLP2017}  &  0\%  & 29\%  &  0\% & 43\%  & & 59.1 & 42.9 & 60.3 & 33.8 \\
    & + Definition LSTM \cite{embeddings-on-fly:2017} &  0\% & 25\%  & 0\%  & 39\% & & 58.6 & 41.8 & 59.4 & 31.7  \\
    & + SemLand \cite{pilehvar-collier:2017:EACLshort} & 0\% & 29\% & 0\% & 43\% & & \bf \underline{60.5}  & \bf \underline{43.4}  & \bf \underline{61.7}  & 34.3  \\
    \midrule
    \multicolumn{2}{l}{\textit{Glove Common Crawl - cased (300d)}} & 1\% & 29\% & 2\% & 44\% & & 44.0 & 33.0 & 45.1 & 27.3\\
    & + Mimick \cite{pinter-guthrie-eisenstein:2017:EMNLP2017}  & 0\%  & ~~0\%  & ~~0\%  & ~~0\% & & \bf 44.7 & 23.9 & 45.6 & 29.5  \\
    & + Definition centroid \cite{herbelot-baroni:2017:EMNLP2017}  & ~~0\% & 21\% & 0\% & 35\% & & 43.5  & 35.2 & 45.1 &  31.7  \\
    & + Definition LSTM \cite{embeddings-on-fly:2017} & 0\% & 20\%  & ~~0\%  &  33\%  & & 24.0 & 23.0  & 22.9  & 19.6  \\
    & + SemLand \cite{pilehvar-collier:2017:EACLshort} & 0\% & 21\%  & ~~0\%  & 35\% & &  44.3  & \bf 39.5  & \bf 45.8 & \bf 33.8  \\
    \midrule
    \multicolumn{2}{l}{FastText \cite{fasttext-subword:TACL999}} &  0\%  & ~~3\%  & 0\%  & ~~5\% & & 46.3 & 19.0 & 48.2 & 20.4 \\

    \bottomrule
  \end{tabular}
  }
  \caption{Correlation performance of different rare and unseen word representation techniques on the Stanford RW and \textsc{Card-660} datasets (the best performance in each batch shown in bold; the overall best underlined).}
  \label{tab:rare_embeddings}
\end{table*} 

\subsection{Rare Word Representation Techniques}
\label{sec:rare_rep}

Rare and unseen word representation has been an active field of research during the past few years, with many different techniques proposed.
In this experiment, we evaluate the performance of some of recent models on our dataset.
These techniques can be broadly classified into two categories.
The first group exploits the knowledge encoded for a rare word in external lexical resources (Section \ref{sec:resource-based}), whereas the second induces embeddings for rare words by extending the semantics of its subword units (Section \ref{sec:subword}).

\subsubsection{Resource-based models}
\label{sec:resource-based}

The basic assumption here is that a lexical resource, such as dictionary, provides high coverage for words in a language, even if they are rare.
Resource-based models usually rely on WordNet as their external resource and estimate the embedding for a rare word by exploiting different types of lexical knowledge encoded for it in the resource.
The \textbf{definition centroid} model of \newcite{Angeliki2017-LAZMWM} takes WordNet word glosses (definitions) as semantic clue.
An embedding is induced for an unseen word by averaging the content words' embeddings in its definition.\footnote{The original model is multimodal (text and images). Given that our focus is on texts, we follow \newcite{herbelot-baroni:2017:EMNLP2017} and use the text modality only.}
The \textbf{definition LSTM} strategy of \newcite{embeddings-on-fly:2017} extends the centroid model by encoding the definition using an LSTM network \cite{hochreiter1997long}, in order to better capture the semantics and word order in the definition.
{\bf SemLand} \cite{pilehvar-collier:2017:EACLshort} also uses WordNet, but takes a different approach which benefits from the graph structure of WordNet.
For an unseen word, SemLand extracts the set of its semantically related words from WordNet and induces an embedding for the unseen word by combining pre-trained embeddings for the related words.

\subsubsection{Subword models}
\label{sec:subword}

Resource-based models fall short of inducing embeddings for words that are not covered in the lexical resource.
Subword models alleviate this limitation by breaking the word into its subword \cite{pinter-guthrie-eisenstein:2017:EMNLP2017,fasttext-subword:TACL999} or morphological units \cite{luong-socher-manning:2013,Botha2014,soricut-och:2015} and induce an embedding by composing the information available for these.
{\bf FastText} \cite{fasttext-subword:TACL999} is one of the popular approaches of this type. The model first splits the unseen word into character ngrams (by default, 3- to 6-grams) and then computes the unseen word's embedding as the centroid of the embeddings of these character $n$-grams (which are available as a result of a specific training).
We also report results for {\bf Mimick} \cite{pinter-guthrie-eisenstein:2017:EMNLP2017}, one of the most recent subword models. The technique learns a mapping function from strings to embeddings by training a Bi-LSTM network that encodes character sequences of a word to its pre-trained embedding.

\subsection{Experimental Setup}

We report results for the five techniques discussed in Sections \ref{sec:subword} and \ref{sec:resource-based}.
We used two of the best performing embedding sets, i.e., Glove cased CC and ConceptNet Numberbatch, to train the models (except FastText for which we use the pre-trained WikiNews subword embeddings\footnote{\url{https://fasttext.cc/docs/en/english-vectors.html}}).
In fact, the models were expected to provide improvements over these baseline embeddings by filling their gaps for unseen words.

Mimick was trained with the default parameters,\footnote{\url{https://github.com/yuvalpinter/Mimick/}} except for the hidden units which we set to 100, instead of the original 50, since the target embeddings in our experiments were larger (300d compared to 128d of the original model).
For the Definition LSTM model, the input definitions were represented as sequences of 50d word embeddings, encoded using an LSTM layer of 100 units, and then passed to a dense layer with 300 neurons with linear activation function.
The training was carried out with Mean Squared Error loss and the RMSprop optimizer, for 100 epochs with batch size 64.

\subsection{Experimental Results}

Table \ref{tab:rare_embeddings} reports the performance of different rare word representation techniques.
Both pre-trained embeddings outperform the IAA of RW, with Glove covering 98\% of the pairs.
This severely limits the room for further meaningful experiments on the dataset.
In contrast, on \textsc{Card-660} and similarly to the previous experiment, there are substantial gaps between IAA (cf. Table \ref{tab:iaa}) and the best-performing models: SemLand and Mimick, with the respective figures of 45.5 ($r$) and 53.3 ($\rho$).
These gaps suggest a difficult dataset which can serve future research in subword and rare word representation as a reliable benchmark.

The definition centroid model proves effective, despite its simplicity, whereas the WordNet-based SemLand provides the best results in most of the settings.
Being constrained to the vocabulary of WordNet, the RW dataset does not constitute a challenging benchmark for WordNet-based models, with most of them providing near full coverage. 
However, these techniques are not as effective on our dataset, with the best WordNet-based model still missing around 33\% of the pairs (with Glove pre-trained embeddings).

The \textsc{Card-660} dataset also proves a very difficult benchmark for subword models.
Despite providing near full coverage, these models are unable to consistently improve the pre-trained word embedding baseline.
This would suggest that the simple strategy of backing off to a word's characters might not always provide reliable means of estimating its semantics (e.g., the single-morpheme word \textit{galaxy}, or the exocentric compound \textit{honeymoon}).  
The results encourage further research on a more semantically-oriented handling of subwords, through learning more effective splitting and composition techniques.

\section{Conclusions}
\label{sec:conclusions}

Thanks to a carefully designed procedure and an expert-based curation, \textsc{Card-660} provides multiple advantages over existing benchmarks, including a very high IAA (average pairwise correlation of around 0.90).
A series of experiments was carried out on the dataset, leading to two main conclusions: (1) the dataset proved a very challenging benchmark, with the best pre-trained embedding model still missing around 40\% of the word pairs and the best rare word representation model hardly crossing into 40.0s (correlation performance); and (2) knowledge-based models are not enough to provide high coverage whereas subword models, which provide near-full coverage, are not semantically as effective.
The significant gap between state of the art and IAA (around 50.0) encourages future research to take this dataset as a challenging, yet reliable, evaluation benchmark.

\section*{Acknowledgments}
We gratefully acknowledge the funding support of EPSRC (N.
Collier and D. Kartsaklis - Grant No. EP/M005089/1) and MRC (M. T. Pilehvar) Grant No. MR/M025160/1 for PheneBank.
We would also like to thank Andreas Chatzistergiou, Costanza Conforti, Gamal Crichton, Milan Gritta, and Ehsan Shareghi for their contribution in creating the dataset.

\bibliography{emnlp2018}

\begin{thebibliography}{34}
\expandafter\ifx\csname natexlab\endcsname\relax\def\natexlab#1{#1}\fi

\bibitem[{Bahdanau et~al.(2017)Bahdanau, Bosc, Jastrzebski, Grefenstette,
  Vincent, and Bengio}]{embeddings-on-fly:2017}
Dzmitry Bahdanau, Tom Bosc, Stanislaw Jastrzebski, Edward Grefenstette, Pascal
  Vincent, and Yoshua Bengio. 2017.
\newblock \href {http://arxiv.org/abs/1706.00286} {Learning to compute word
  embeddings on the fly}.
\newblock \emph{CoRR}, abs/1706.00286.

\bibitem[{Baroni et~al.(2009)Baroni, Bernardini, Ferraresi, and
  Zanchetta}]{Baroni2009Wacky}
Marco Baroni, Silvia Bernardini, Adriano Ferraresi, and Eros Zanchetta. 2009.
\newblock The {WaCky} wide web: a collection of very large linguistically
  processed web-crawled corpora.
\newblock \emph{Language Resources and Evaluation}, 43(3):209--226.

\bibitem[{Blitzer et~al.(2007)Blitzer, Dredze, and Pereira}]{P07-1056}
John Blitzer, Mark Dredze, and Fernando Pereira. 2007.
\newblock Biographies, bollywood, boom-boxes and blenders: Domain adaptation
  for sentiment classification.
\newblock In \emph{Proceedings of the 45th Annual Meeting of the Association of
  Computational Linguistics}, pages 440--447. Association for Computational
  Linguistics.

\bibitem[{Bojanowski et~al.(2017)Bojanowski, Grave, Joulin, and
  Mikolov}]{fasttext-subword:TACL999}
Piotr Bojanowski, Edouard Grave, Armand Joulin, and Tomas Mikolov. 2017.
\newblock Enriching word vectors with subword information.
\newblock \emph{Transactions of the Association for Computational Linguistics},
  5:135--146.

\bibitem[{Bollacker et~al.(2008)Bollacker, Evans, Paritosh, Sturge, and
  Taylor}]{Bollackeretal:2008}
Kurt Bollacker, Colin Evans, Praveen Paritosh, Tim Sturge, and Jamie Taylor.
  2008.
\newblock Freebase: A collaboratively created graph database for structuring
  human knowledge.
\newblock In \emph{Proceedings of the 2008 ACM SIGMOD International Conference
  on Management of Data}, pages 1247--1250, Vancouver, Canada.

\bibitem[{Botha and Blunsom(2014)}]{Botha2014}
Jan~A. Botha and Phil Blunsom. 2014.
\newblock {Compositional Morphology for Word Representations and Language
  Modelling}.
\newblock In \emph{Proceedings of ICML}, pages 1899--1907, Beijing, China.

\bibitem[{Bruni et~al.(2014)Bruni, Tran, and Baroni}]{Men3k:2014}
Elia Bruni, Nam~Khanh Tran, and Marco Baroni. 2014.
\newblock Multimodal distributional semantics.
\newblock \emph{Journal of Artificial Intelligence Research}, 49(1):1--47.

\bibitem[{Camacho-Collados et~al.(2017)Camacho-Collados, Pilehvar, Collier, and
  Navigli}]{camachocollados-EtAl:2017:SemEval}
Jose Camacho-Collados, Mohammad~Taher Pilehvar, Nigel Collier, and Roberto
  Navigli. 2017.
\newblock Semeval-2017 task 2: Multilingual and cross-lingual semantic word
  similarity.
\newblock In \emph{Proceedings of the 11th International Workshop on Semantic
  Evaluation (SemEval-2017)}, pages 15--26, Vancouver, Canada.

\bibitem[{Fellbaum(1998)}]{Fellbaum:98}
Christiane Fellbaum, editor. 1998.
\newblock \emph{{W}ord{N}et: An Electronic Database}.
\newblock MIT Press, Cambridge, MA.

\bibitem[{Finkelstein et~al.(2002)Finkelstein, Evgenly, Yossi, Ehud, Zach,
  Gadi, and Eytan}]{Finkelsteinetal:2002}
Lev Finkelstein, Gabrilovich Evgenly, Matias Yossi, Rivlin Ehud, Solan Zach,
  Wolfman Gadi, and Ruppin Eytan. 2002.
\newblock Placing search in context: The concept revisited.
\newblock \emph{ACM Transactions of Information Systems}, 20(1):116--131.

\bibitem[{Gerz et~al.(2016)Gerz, Vuli{\'{c}}, Hill, Reichart, and
  Korhonen}]{Gerzetal:2016}
Daniela Gerz, Ivan Vuli{\'{c}}, Felix Hill, Roi Reichart, and Anna Korhonen.
  2016.
\newblock Simverb-3500: A large-scale evaluation set of verb similarity.
\newblock In \emph{Proceedings of EMNLP}, pages 2173--2182.

\bibitem[{Greene and Cunningham(2006)}]{greene06icml}
Derek Greene and P\'{a}draig Cunningham. 2006.
\newblock Practical solutions to the problem of diagonal dominance in kernel
  document clustering.
\newblock In \emph{Proc. 23rd International Conference on Machine learning
  (ICML'06)}, pages 377--384. ACM Press.

\bibitem[{Herbelot and Baroni(2017)}]{herbelot-baroni:2017:EMNLP2017}
Aur\'{e}lie Herbelot and Marco Baroni. 2017.
\newblock High-risk learning: acquiring new word vectors from tiny data.
\newblock In \emph{Proceedings of the 2017 Conference on Empirical Methods in
  Natural Language Processing}, pages 304--309, Copenhagen, Denmark.

\bibitem[{Hill et~al.(2015)Hill, Reichart, and Korhonen}]{Hilletal:2015}
Felix Hill, Roi Reichart, and Anna Korhonen. 2015.
\newblock {SimLex-999}: Evaluating semantic models with (genuine) similarity
  estimation.
\newblock \emph{Computational Linguistics}, 41(4):665--695.

\bibitem[{Hochreiter and Schmidhuber(1997)}]{hochreiter1997long}
Sepp Hochreiter and J{\"u}rgen Schmidhuber. 1997.
\newblock Long short-term memory.
\newblock \emph{Neural computation}, 9(8):1735--1780.

\bibitem[{Kim et~al.(2004)Kim, Ohta, Tsuruoka, Tateisi, and
  Collier}]{Kim:2004:IBR}
Jin-Dong Kim, Tomoko Ohta, Yoshimasa Tsuruoka, Yuka Tateisi, and Nigel Collier.
  2004.
\newblock Introduction to the bio-entity recognition task at {JNLPBA}.
\newblock In \emph{Proceedings of the International Joint Workshop on Natural
  Language Processing in Biomedicine and Its Applications}, pages 70--75,
  Stroudsburg, PA, USA. Association for Computational Linguistics.

\bibitem[{Koehn(2005)}]{koehn2005epc}
Philipp Koehn. 2005.
\newblock {Europarl: A Parallel Corpus for Statistical Machine Translation}.
\newblock In \emph{{Conference Proceedings: the tenth Machine Translation
  Summit}}, pages 79--86, Phuket, Thailand. AAMT, AAMT.

\bibitem[{Lazaridou et~al.(2017)Lazaridou, Marelli, and
  Baroni}]{Angeliki2017-LAZMWM}
Angeliki Lazaridou, Marco Marelli, and Marco Baroni. 2017.
\newblock Multimodal word meaning induction from minimal exposure to natural
  text.
\newblock \emph{Cognitive Science}, 41(S4):677--705.

\bibitem[{Leviant and Reichart(2015)}]{DBLP:journals/corr/LeviantR15}
Ira Leviant and Roi Reichart. 2015.
\newblock \href {http://arxiv.org/abs/1508.00106} {Judgment language matters:
  Multilingual vector space models for judgment language aware lexical
  semantics}.
\newblock \emph{CoRR}, abs/1508.00106.

\bibitem[{Levy and Goldberg(2014)}]{levy-goldberg:2014:P14-2}
Omer Levy and Yoav Goldberg. 2014.
\newblock Dependency-based word embeddings.
\newblock In \emph{Proceedings of the 52nd Annual Meeting of the Association
  for Computational Linguistics (Volume 2: Short Papers)}, pages 302--308,
  Baltimore, Maryland. Association for Computational Linguistics.

\bibitem[{Ling et~al.(2017)Ling, Yogatama, Dyer, and
  Blunsom}]{ling-EtAl:2017:Long}
Wang Ling, Dani Yogatama, Chris Dyer, and Phil Blunsom. 2017.
\newblock Program induction by rationale generation: Learning to solve and
  explain algebraic word problems.
\newblock In \emph{Proceedings of the 55th Annual Meeting of the Association
  for Computational Linguistics (Volume 1: Long Papers)}, pages 158--167,
  Vancouver, Canada. Association for Computational Linguistics.

\bibitem[{Luong et~al.(2013)Luong, Socher, and
  Manning}]{luong-socher-manning:2013}
Thang Luong, Richard Socher, and Christopher Manning. 2013.
\newblock Better word representations with recursive neural networks for
  morphology.
\newblock In \emph{Proceedings of CoNLL}, pages 104--113, Sofia, Bulgaria.

\bibitem[{Maas et~al.(2011)Maas, Daly, Pham, Huang, Ng, and
  Potts}]{maas-EtAl:2011:ACL-HLT2011}
Andrew~L. Maas, Raymond~E. Daly, Peter~T. Pham, Dan Huang, Andrew~Y. Ng, and
  Christopher Potts. 2011.
\newblock Learning word vectors for sentiment analysis.
\newblock In \emph{Proceedings of ACL-HLT}, pages 142--150, Portland, Oregon,
  USA.

\bibitem[{Mikolov et~al.(2013)Mikolov, Chen, Corrado, and
  Dean}]{Mikolovetal:2013}
Tomas Mikolov, Kai Chen, Greg Corrado, and Jeffrey Dean. 2013.
\newblock Efficient estimation of word representations in vector space.
\newblock In \emph{Workshop at ICLR}, Scottsdale, Arizona.

\bibitem[{Pennington et~al.(2014)Pennington, Socher, and
  Manning}]{Penningtonetal:2014}
Jeffrey Pennington, Richard Socher, and Christopher Manning. 2014.
\newblock Glove: Global vectors for word representation.
\newblock In \emph{Proceedings of EMNLP 2014}, pages 1532--1543, Doha, Qatar.

\bibitem[{Pilehvar and Collier(2017)}]{pilehvar-collier:2017:EACLshort}
Mohammad~Taher Pilehvar and Nigel Collier. 2017.
\newblock Inducing embeddings for rare and unseen words by leveraging lexical
  resources.
\newblock In \emph{Proceedings of the 15th Conference of the European Chapter
  of the Association for Computational Linguistics: Volume 2, Short Papers},
  pages 388--393, Valencia, Spain.

\bibitem[{Pinter et~al.(2017)Pinter, Guthrie, and
  Eisenstein}]{pinter-guthrie-eisenstein:2017:EMNLP2017}
Yuval Pinter, Robert Guthrie, and Jacob Eisenstein. 2017.
\newblock Mimicking word embeddings using subword rnns.
\newblock In \emph{Proceedings of the 2017 Conference on Empirical Methods in
  Natural Language Processing}, pages 102--112, Copenhagen, Denmark.

\bibitem[{Rajpurkar et~al.(2016)Rajpurkar, Zhang, Lopyrev, and
  Liang}]{rajpurkar-EtAl:2016:EMNLP2016}
Pranav Rajpurkar, Jian Zhang, Konstantin Lopyrev, and Percy Liang. 2016.
\newblock {SQuAD}: 100,000+ questions for machine comprehension of text.
\newblock In \emph{Proceedings of the 2016 Conference on Empirical Methods in
  Natural Language Processing}, pages 2383--2392, Austin, Texas. Association
  for Computational Linguistics.

\bibitem[{Rubenstein and Goodenough(1965)}]{RG65:1965}
Herbert Rubenstein and John~B. Goodenough. 1965.
\newblock Contextual correlates of synonymy.
\newblock \emph{Communications of the ACM}, 8(10):627--633.

\bibitem[{Salle et~al.(2016)Salle, Villavicencio, and
  Idiart}]{salle-villavicencio-idiart:2016:P16-2}
Alexandre Salle, Aline Villavicencio, and Marco Idiart. 2016.
\newblock Matrix factorization using window sampling and negative sampling for
  improved word representations.
\newblock In \emph{Proceedings of the 54th Annual Meeting of the Association
  for Computational Linguistics (Volume 2: Short Papers)}, pages 419--424,
  Berlin, Germany. Association for Computational Linguistics.

\bibitem[{Sergienya and Sch{\"u}tze(2015)}]{SergienyaSchutze:2015}
Irina Sergienya and Hinrich Sch{\"u}tze. 2015.
\newblock Learning better embeddings for rare words using distributional
  representations.
\newblock In \emph{Proceedings of EMNLP}, pages 280--285.

\bibitem[{Soricut and Och(2015)}]{soricut-och:2015}
Radu Soricut and Franz Och. 2015.
\newblock Unsupervised morphology induction using word embeddings.
\newblock In \emph{Proceedings of NAACL-HLT}, pages 1627--1637, Denver,
  Colorado.

\bibitem[{Speer et~al.(2017)Speer, Chin, and Havasi}]{speer2017conceptnet}
Robert Speer, Joshua Chin, and Catherine Havasi. 2017.
\newblock {ConceptNet} 5.5: An open multilingual graph of general knowledge.
\newblock In \emph{AAAI Conference on Artificial Intelligence}, pages
  4444--4451.

\bibitem[{Yang and Powers(2005)}]{Yang06verbsimilarity}
Dongqiang Yang and David M.~W. Powers. 2005.
\newblock Measuring semantic similarity in the taxonomy of {WordNet}.
\newblock In \emph{Proceedings of the Twenty-eighth Australasian Conference on
  Computer Science}, volume~38, pages 315--322, Newcastle, Australia.

\end{thebibliography}
\bibliographystyle{acl_natbib}

\appendix

\end{document}